%% file: main.tex
\documentclass[lettersize,journal]{IEEEtran} 
\IEEEoverridecommandlockouts                              % This command is only
\usepackage{graphicx}
\usepackage{amsmath,amssymb}
\usepackage{cite}
\usepackage{float}
\usepackage[caption=false,font=footnotesize]{subfig}
\usepackage{algorithm}
\usepackage{algorithmic}
\usepackage{color}
\definecolor{review_color}{rgb}{0,0,0}

\graphicspath{{img/},{fig/}}

\begin{document}

\input{src}

%%%%%%%%%%%%%%%%%%%%%%%%%%%%%%%%%%%%%%%%%%%%%%%%%%%%%%%%%%%%%%%%%%%%%%%%%%%%%

%************************************************************************
% References                                                                     
%************************************************************************

\bibliographystyle{IEEEtran}
\bibliography{main}

\end{document}

%% file: src.tex
% \title{\LARGE \bf
% A Two-Speed Actuator for Robotics with Distinctively Different Gear Ratios and Fast Seamless Gear Shifting
% }
% A two-speed actuator for robotics with fast seamless gear shifting
\title{\LARGE \bf
A Two-Speed Actuator for Robotics with Fast Seamless Gear Shifting
}

%%%%%%%%%%%%%%%%%%%%%%%%%%%%%%%%%%%%%%%%%%%%%%%%%%%%%%%%%%%%%%%%%%%%%%%%%%%%%

% \author{Alexandre Girard,~\IEEEmembership{Student Member,~IEEE,} and H. Harry Asada,~\IEEEmembership{Member,~IEEE}% <-this % stops a space
\author{Alexandre Girard$^1$ and H. Harry Asada$^2$% <-this % stops a space
\thanks{This work was supported by The Boeing Company. }%
\thanks{$^1$ A. Girard is with the Department of Mechanical Engineering, Universite de Sherbrooke, Qc, Canada.} 
\thanks{$^2$ H. H. Asada is with Department of Mechanical Engineering, Massachusetts Institute of Technology, Cambridge, MA, USA.}%
\thanks{$^3$ \textcopyright IEEE. Personal use of this material is permitted. Permission from IEEE must be obtained for all other uses, in any current or future media, including reprinting/republishing this material for advertising or promotional purposes, creating new collective works, for resale or redistribution to servers or lists, or reuse of any copyrighted component of this work in other works. DOI:10.1109/IROS.2015.7354047}
}

\markboth{2015 IEEE/RSJ International Conference on Intelligent Robots and Systems (IROS), Preprint version. DOI:10.1109/IROS.2015.7354047$^3$}{}

\maketitle
%\thispagestyle{empty}
%\pagestyle{empty}

%%%%%%%%%%%%%%%%%%%%%%%%%%%%%%%%%%%%%%%%%%%%%%%%%%%%%%%%%%%%%%%%%%%%%%%%%%%%%%%%
%%%%%%%%%%%%%%%%%%%%%%%%%%%%%%%%%%%%%%%%%%%%%%%%%%%%%%%%%%%%%%%%%%%%%%%%%%%%%%%%
\begin{abstract}
This paper present a novel dual-speed actuator adapted to robotics. In many applications, robots have to bear large loads while moving slowly and also have to move quickly through the air with almost no load. This lead to conflicting requirements for their actuators. Multiple gear ratios address this issue by allowing an effective use of power over a wide range of torque-speed load conditions. Furthermore, very different gear ratios also lead to drastic changes of the intrinsic impedance, enabling a non-back-drivable mode for stiff position control and a back-drivable mode for force control. The proposed actuator consists of two electric motors coupled to a differential; one has a large gear ratio while the other is almost direct-drive and equipped with a brake. During the high-force mode the brake is locked, only one motor is used, and the actuator behaves like a regular highly-geared servo-motor. During the high-speed mode the brake is open, both motor are used at the same time, and the actuator behaves like a direct drive motor. A dynamic model is developed and novel controllers are proposed for synergic use of both motors. The redundancy of motors is exploited for maintaining full control of the output during mode transitions, allowing for fast and seamless switching even when interacting with unknown environments. Results are demonstrated with a proof-of-concept linear actuator.
\end{abstract}

%\begin{IEEEkeywords}
%mobile manufacturing robots, actuators, two-speed, dual mode, dual motors, micro-macro actuators
%\end{IEEEkeywords}

%%%%%%%%%%%%%%%%%%%%%%%%%%%%%%%%%%%%%%%%%%%%%%%%%%%%%%%%%%%%%%%%%%%%%%%%%%%%%%%
\section{INTRODUCTION}
\begin{figure}[htbp]
	\centering
		\includegraphics[width=0.45\textwidth]{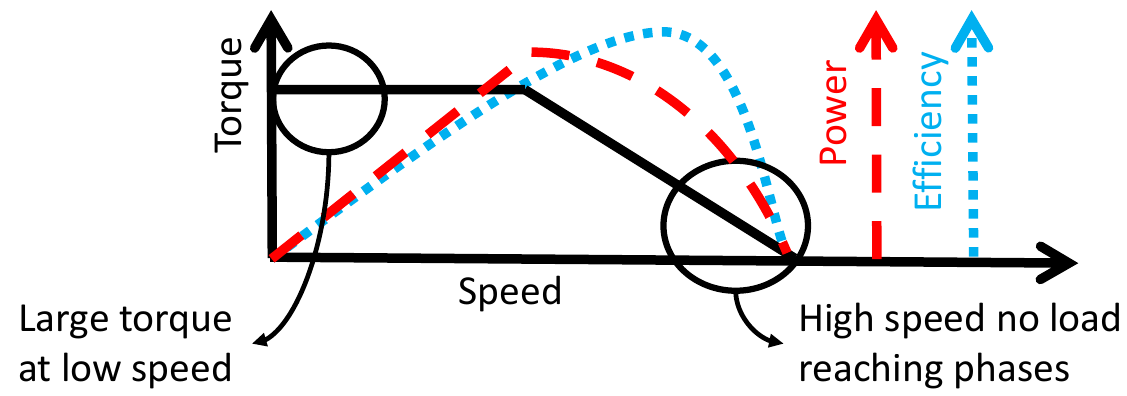}
	\caption{Limitations of EM motors for extremum torque-speed operations}
	\label{fig:speedissue}
	\vspace{-10pt}
\end{figure}

In many robotic systems, including legged robots and wearable robots, actuators are often required to operate in distinctively different torque-speed load conditions. A legged robot, for example, has to move its leg forward quickly through the air and, once touching the ground, it has to bear a large load \cite{hirose_study_1984}. These two operating conditions, high speed at low torque vs. high torque at low speed, are often an order of magnitude different, while the required output power is similarly low. When the torque-speed load conditions do not vary significantly, a single gear ratio can be picked so that the actuator always operate under nearly optimal conditions. For distinctively different torque-speed conditions, the actuator will be far for its optimal operating conditions with a gear ratio picked from the middle ground. As illustrated on Fig. \ref{fig:speedissue} for a typical electromagnetic (EM) actuator, extremum torque-speed conditions are not optimal in term of efficiency and power output. This often leads to the use of oversized and inefficient actuators, which is inhibitory particularly for mobile robots.

%\begin{figure}[h]
	%\centering
		%\includegraphics[width=0.40\textwidth]{../img/proto_linear_2.jpg}
	%\label{fig:proto_linear_2}
	%\caption{Two-Speed Actuator Prototype}
%\end{figure}

%

Automobiles with internal combustion (IC) engines use transmissions with multiple gear ratios to match torque-speed conditions. IC engines have a very narrow speed range in which they can effectively deliver power; a transmission with multiple gear ratios is a necessity for the engine to work effectively for a wide range of output speed. EM motors are more flexible than IC engines, but still far from ideal sources. EM motors cannot output high power at low speed because of thermal dissipation and magnetic flux limits related to material properties; are limited in speed by the supply tension and others; and are very inefficient when producing large forces at low speed \cite{hollerbach_comparative_1992}. In robotic, since it is often the extremums, i.e. maximum torque and speed, that determine the actuator design instead of the power requirement, much can be gained with multiple gear ratios.

It will be a significant breakthrough if a type of multiple speed transmission can be used effectively in robotics. The use of multiple reduction ratios would enable actuators to meet diverse speed-torque load conditions. Even a small, lightweight actuator can generate large torques and move at high speed if equipped with both large and small gear ratio. Moreover, multiple speed transmission can allow an actuator to work closer to its optimal operating conditions, improving overall efficiency significantly. Furthermore, gear shifting significantly changes the intrinsic impedance of an actuator. Since the impedance is proportional to the square of gear ratio, the impedance may vary in a range that is several order-of-magnitude different. The actuator may be made back-drivable while using its small reduction ratio, an important property in many applications where the robot physically interacts with the environment \cite{hogan_impedance_2004}. Also the same actuator may be made non-back-drivable while using its large reduction ratio, allowing the actuator to support loads without consuming energy and enabling high-stiffness position control.

Despite the desirable features, gear shifting is more technically challenging in robotics applications than in vehicle applications. For powertrains, the load is mostly a large inertia, while for robots, the loads may exhibit a rich range of dynamics including spring-like and damper-like loads. Hence, unlike vehicle applications, leaving the load free momentarily during transitions (from one gear ratio to another) is not acceptable in the context of robotics; the actuator must be engaged with the load at all time to maintain control of the output. Hence, an effective gear shifting methodology is necessary for seamless transitions under diverse load conditions.

This paper presents a novel two-speed actuator that meets the requirements of the robotics context. The system consist of a highly geared EM motor and a direct drive EM motor equipped with a locking brake, both engaged with a 3-port gearbox. In high-force mode the actuator is non-back-drivable and can produce large torque at low speed, while in high-speed mode the actuator is back-drivable and capable of force control and high-speed motion. Moreover, the two motors are coordinated to fully control the output load, while performing fast and seamless mode transitions. In the following, related works are briefly reviewed (section \ref{act}), then the principle of the proposed two-speed actuator is presented (section \ref{dual}). Next, a dynamic model is derived (section \ref{sec:Modeling}) and control algorithms for fast seamless transition and synergistic coordination of the dual motors will be discussed (section \ref{sec:ctl}). The results are demonstrated experimentally with a proof-of-concept linear actuator (section \ref{ev}).

%%%%%%%%%%%%%%%%%%%%%%%%%%%%%%%%%%%%%%%%%%%%%%%%%%%%%%%%%%%%
\section{Related Works}
\label{act}

Traditional robots generally use actuators that behave as displacement-sources because of their high intrinsic impedance. These include geared EM motors and hydraulics cylinders. Using a force sensor, it is possible to control the output force using those actuators, but the bandwidth is rather limited. To guarantee the stability of the force-feedback scheme only half the intrinsic inertia can be canceled \cite{hogan_impedance_2004}. Since 70's, roboticists have been attempting to build actuators that can behave naturally as a force-source such as series-elastic actuators, pneumatic cylinder and air-muscles \cite{hanafusa_stable_1977}\cite{pratt_series_1995}. However, because of the physical limitation of compliant transmission materials, the achievable bandwidth is limited and precise position control is hardly achievable. Another interesting approach is using force-controllable clutches to decouple a geared EM motor with high-impedance \cite{fauteux_dual-differential_2010}. However, those actuators using friction-drive or electro/magnetorehologic fluids clutches have poor efficiency and limited bandwidth and fidelity due to physical limitation and difficulties for accurate control. Direct drive EM actuators are the best force-source actuators with high fidelity, high bandwidth. However, the very low force density and low efficiency at low speeds make them impractical for mobile robot applications \cite{hollerbach_comparative_1992}. On the other hand, actuators with non-back-drivable mechanisms have the advantage for pure position control tasks and they can bear very large load without any power consumption.

Since both small and large intrinsic impedances are advantageous in different scenario, several group have developed variable intrinsic impedance actuators. Novel design concepts, such as those using a variable stiffness spring \cite{tonietti_design_2005} or antagonist non-linear devices \cite{koganezawa_antagonistic_2006}, can alter the stiffness continually, yet the variable range is rather limited. A promising approach using a series-compliance that can be locked with a brake vary the impedance in a broader range \cite{leach_linear_2012}. Another concept is a dual actuator systems using EM motors in a serial configuration, one controlling the position and the other the impedance \cite{kim_serial-type_2010}. Furthermore, so-called macro-micro actuators, can improve the bandwidth of force-source type of actuators by exploiting the high-bandwidth of a small actuator in parallel, allowing for wider-range impedance control and improved position control\cite{morrell_parallel-coupled_1998}.

While the actuator work in robotics have been focused on impedance and bandwidth issues, in the powertrain field the torque-speed matching issue is predominant, since power density and efficiency are critical for mobile systems. The idea of using multiple gear ratios with electric motors has been explored occasionally, to improve efficiency and power density \cite{mckeegan_antonovs_2011}. A twin motor configuration has been proposed for smooth gear shifting, where each motor shifts at a different timing \cite{bologna_electric_2014}. Also, a dual motor configuration using a planetary coupling and non-back-drivable worm-gears was proposed for a mobile robot powertrain \cite{lee_new_2012}. Multiple speed powertrains provide effective solutions for torque-speed matching, however robots require more distinct gear ratios and a gear shifting methodology adapted to the wide dynamic behaviors of the load.

\begin{table}[tb]
	\centering
		\label{tab:ActuatorTechnologies}
		\begin{tabular}{ c c c c c}
		\hline
		Actuator                  &  Force     & Stiff position   & Ideal op.                        & Holding  \\
		technology                &  control   & control          & speed\footnotemark[1]            & efficiency \\
		\hline\hline
		Geared EM                 & Poor       &  Good        & Low                  & Good \\
		Direct drive EM           & Good       &  Fair        & High                 & Poor \\
		Series elastic             & Fair       &  Poor        & Low\footnotemark[2]  & Good \\
		Micro-Macro               & Good       &  Fair        & Low\footnotemark[2]  & Good \\
		Serial Dual Motors        & Good       &  Fair        & High                 & Poor \\
		Clutch actuators          & Fair       &  Poor        & Low\footnotemark[2]  & Poor \\
		Pneumatic                 & Fair       &  Poor        & High                 & Good \\
		Hydraulic                 & Poor       &  Good        & Low                  & Good \\
		\hline			\\
		\multicolumn{5}{l}{ $^1$ Operating speed at which the actuator is powerful and efficient.} \\
		\multicolumn{5}{l}{ $^2$ Assuming the primary actuator is a geared EM motor. }
		\end{tabular}
			\caption{Actuator technologies for robotics}
		\vspace{-25pt}
\end{table}

Table I summarizes the comparison of the actuators discussed above. Note that the actuators developed for good force control performance have lost many advantages of traditional high impedance actuators. The proposed dual-speed actuator with two distinctively different gear ratios is an alternative approach that combines the advantages of both types of actuators. Compared to the existing variable impedance actuators, the new design can achieve order-of-magnitude different impedance instead of continuous small variations. Moreover, the proposed dual-speed design aims to improve available power and efficiency over a wide range of operating speed, which has rarely been addressed in the robotics literature. All actuators listed in Table \ref{tab:ActuatorTechnologies} are powerful and efficient for either low or high output speed alone. Furthermore, unlike the powertrains of automobiles, the proposed concept has distinctively different reduction ratio and the output load is always fully under control even during the transitions.

%%%%%%%%%%%%%%%%%%%%%%%%%%%%%%%%%%%%%%%%%%%%%%%%%%%%%%%%%%%%%%%

\section{Dual-speed dual-motor actuator principle}
\label{dual}

\begin{figure}[htb]
	\centering
		\includegraphics[width=0.40\textwidth]{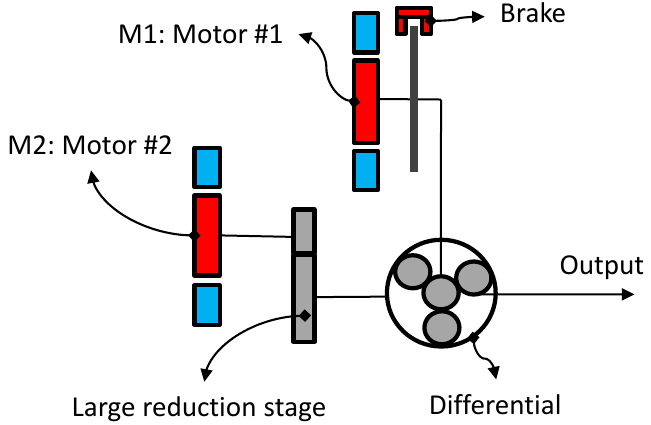}
	\caption{DSDM actuator concept}
	\label{fig:dualmotorconcept}
\end{figure}

The new design concept, referred to as a Dual-Speed Dual-Motor (DSDM) actuator, consists of a direct drive motor (M1) equipped with a locking brake and an geared EM motor (M2) with a large reduction ratio coupled to the same output through a differential, see Fig. \ref{fig:dualmotorconcept}. The differential can be viewed as a series-type junction where the speeds add up and the force is shared. 

\begin{figure}[htb]
	\centering
		\includegraphics[width=0.40\textwidth]{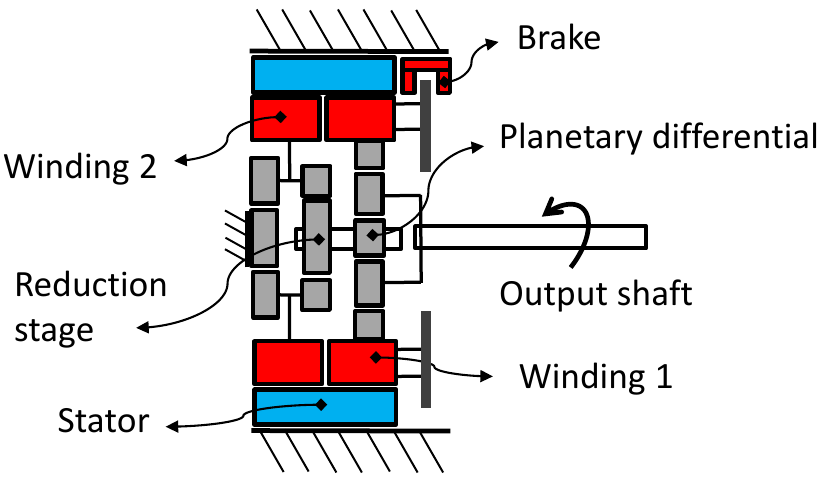}
	\caption{Possible architecture of an integrated DSDM concept}
	\label{fig:embedded}
\end{figure}

The envisioned implementation of the DSDM concept is to embed all the components into a single compact unit, as illustrated by Fig. \ref{fig:embedded}. A lot of weight and space could be saved by combining the reduction and the differential gearing and having all the components inside a single housing. 

\begin{figure}[htb]
	\centering
		\includegraphics[width=0.40\textwidth]{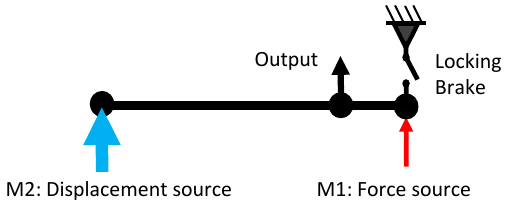}
	\caption{Dual input system}
	\label{fig:lever}
\end{figure}

\begin{figure}[htb]
        \centering
				\subfloat[High force mode]{
        \includegraphics[width=0.22\textwidth]{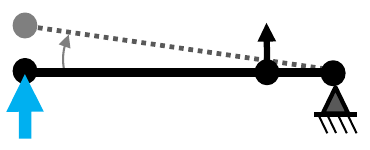}
				\label{fig:HF}}
        \subfloat[High speed mode]{
				\includegraphics[width=0.22\textwidth]{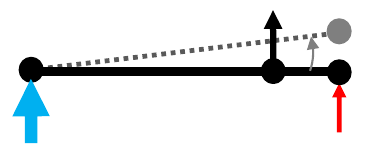}
				\label{fig:HS}}
        \caption{Two modes of operation}\label{fig:opmode}
\end{figure}

The DSDM can be used in two modes, high-force mode when the brake is closed and high-speed mode when the brake is open. The result is like having two very different reduction ratio you can choose from during operation. Fig. \ref{fig:lever} conceptually illustrates the principle with a leverage analogy, M1 acts like a force source connected almost directly to the output and M2 acts like a displacement source with a large lever arm relative to the output. 
During the high-force mode, see Fig. \ref{fig:HF}, the brake is closed and M2 drives the output with a large mechanical advantage. The result is a low-speed displacement-source type of actuation like a geared EM motor. During the high-speed mode, see Fig. \ref{fig:HS}, M1 drive the output almost directly, creating a high-speed force-source actuator like a direct drive EM motor. Additionally, both motors can be used simultaneously to drive the output even faster.

\begin{figure}[htb]
	\centering
		\includegraphics[width=0.45\textwidth]{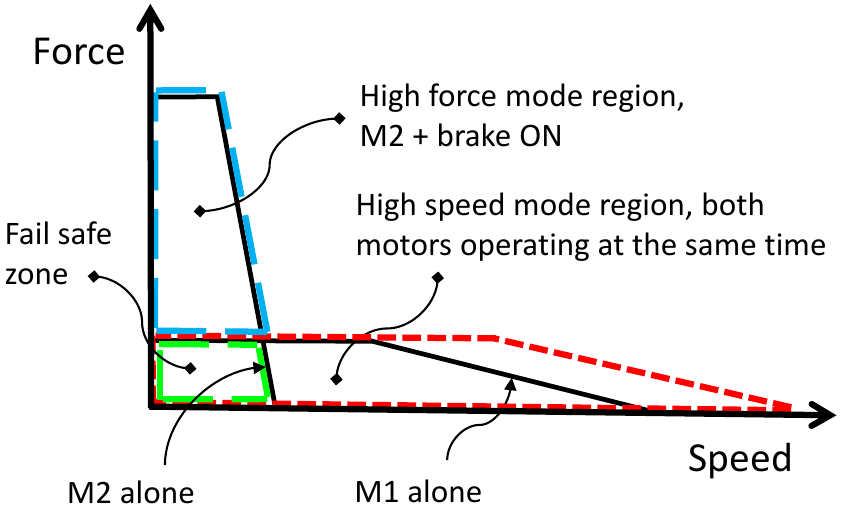}
	\caption{DSDM actuator operation region, with a difference between M1 and M2 gearing ratio of only 4 for illustration purposes }
	\label{fig:torquespeed}
\end{figure}

Fig. \ref{fig:torquespeed} illustrates the operating range of the DSDM actuator plotted on the standard torque-speed plane. The high-force mode region is determined by the performance of M2 alone, since M1 is locked. The high-speed mode region can exceed the performance of M1 alone, as M2 can be used simultaneously to increase the output speed. The fail safe zone indicate the guaranteed performance of the DSDM actuator in case of failure in either motor. 

%%%%%%%%%%%%%%%%%%%%%%%%%%%%%%%%%%%%%%%%%%%%%%%%%%%%%%%%%
\subsection{Weight advantage}
\label{sec:WeightAdvantage}

\begin{figure}[htb]
        \centering
				\subfloat[One motor solution]{
				\includegraphics[width=0.22\textwidth]{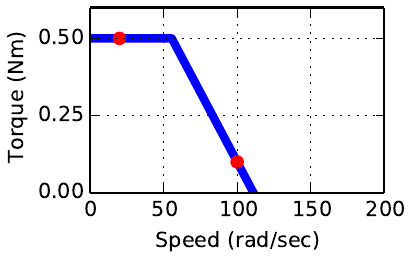}
				\label{fig:s1}}
        \subfloat[DSDM solution]{
				\includegraphics[width=0.22\textwidth]{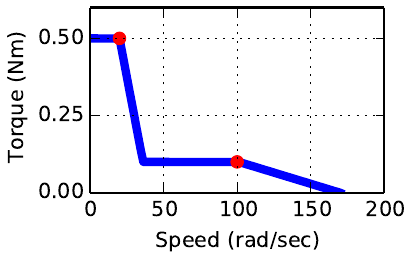}
				\label{fig:s2}}
        \caption{Case study of two actuator solution for two 10 W operating points }\label{fig:solutions}
\end{figure}

\begin{figure}[htb]
	\centering
		\includegraphics[width=0.45\textwidth]{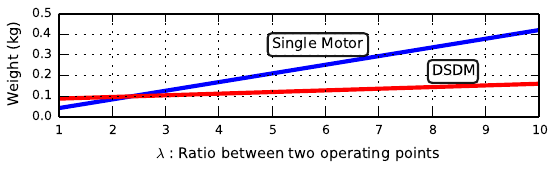}
	\caption{Weight of a single motor compared to the DSDM concept for two 10 W operating points at different speeds $w_1=100$ rad/sec, $w_2 = w_1 / \lambda$}
	\label{fig:1vs2}
\end{figure}

A DSDM actuator will be lighter than a single motor for applications with a wide range of operating speed. Suppose that an actuator must generate 10 W output power at two operating points: 0.5 Nm of torque at a speed of 20 rad/sec and 0.1 Nm at 100 rad/sec. A single EM motor that satisfies these requirements at both operating points tends to be oversized in terms of power, to reach both operating points, see Fig. \ref{fig:s1}. A DSDM actuator can reach the same operating points using two smaller motors with appropriate gear ratios, see Fig. \ref{fig:s2}. On the other hand the DSDM actuator uses more components: two motors instead of one, more gearing and an additional brake. The DSDM concept pays-off when the difference in speed between two required operating points becomes larger.  Fig. \ref{fig:1vs2} shows the estimated weight of actuators in relation to the ratio of operating speeds ($\lambda=\frac{w_1}{w_2}$), while the required power output is kept at 10 W. The actuator weight is computed assuming that the mass of each component is proportional to its maximum output torque, with values taken from commercially available components in the 10 - 100 watts range: 2 kg/Nm for motors, 0.1 kg/Nm for gearboxes and differentials and 0.2 kg/Nm for brakes \cite{maxon_motor_usa}. As shown in Fig. \ref{fig:1vs2}, the DSDM concept becomes advantageous when there is a large speed difference between the operating points. This is because only the gearbox and brake need to be scaled up for the DSDM actuator to meet the high torque requirement of the low-speed operating point, while the motor size must be increased for the single motor solution.

%%%%%%%%%%%%%%%%%%%%%%%%%%%%%%%%%%%%%%%%%%%%%%%%%%%%%%%%%
\subsection{Efficiency advantage}
\label{sec:eff}

Since EM motor are not efficient when producing large torque at low speeds, the DSDM concept can be very advantageous in term of efficiency. The power loss $P_{loss}$ in an EM motor is proportional to the square of the motor output torque $\tau_m$. Hence, for a given output torque $\tau_{out}$, the motor with the larger gear ratio $R$ will be more efficient:
\begin{align}
	P_{loss} = r I^2 = \frac{r}{k_m^2} \tau_m^2 = \frac{r \tau_{out}^2}{k_m^2 R^2} 
	\label{eq:loss}
\end{align}
where $r$ is the winding resistance and $k_m$ is the motor torque constant. Taking the same torque-speed requirements used for analyzing the weight advantage (see Fig. \ref{fig:1vs2}), the single motor would have power loss $\lambda^2$ times greater than the DSDM at the low speed operating point.

%%%%%%%%%%%%%%%%%%%%%%%%%%%%%%%%%%%%%%%%%%%%%%%%%%%%%%%%%
\section{Modeling}
\label{sec:Modeling}
%%%%%%%%%%%%%%%%%%%%%%%%%%%%%%%%%%%%%%%%%%%%%%%%%%%%%%%%%

\subsection{3-ports planetary gear junction}
If a planetary gear box is used to implement the serial junction with the planet carrier connected to the output, M1 to the sun gear and M2 to the ring gear, the kinematic relation of the system is given by
\begin{align}
	w_o = 
	\underbrace{ \left[ 
	\frac{ 1 }{N+1}
	\right] }_{  1/R_1  }
	w_1 + 
	\underbrace{ \left[ 
	\frac{ N  }{r_2 (N+1)}
	\right] }_{  1/R_2  }
	w_2
\label{eq:kinematic}
\end{align}
where $r_2$ is the additional reduction of M2, $N$ is the ratio of gear teeth of the ring gear over the sun gear, and $w_o$, $w_1$ and $w_2$ are angular velocities of the output shaft (port $o$), M1 input shaft (port $1$) and M2 input shaft (port $2$). Assuming the planet gears are massless, the torque relation of the system is given by:
\begin{align}
	- \tau_o =
	\underbrace{ \left[ 
	N+1
	\right] }_{ R_1  }
	\tau_1 = 
	\underbrace{ \left[ 
	\frac{r_2(N+1)}{N}
	\right] }_{ R_2  }
	\tau_2
	\label{eq:torque}
\end{align}
Hence, the 3-ports planetary coupling can be interpreted as a 0-junction, in the bond graph terminology, with different mechanical advantages ($R_1$ and $R_2$) on each input ports. 

%%%%%%%%%%%%%%%%%%%%%%%%%%%%%%%%%%%%%%%%%%%%%%%%%%%%%%%%%
\subsection{Dynamics}
\label{sec:dyn}

\begin{figure}[htb]
	\centering
		\includegraphics[width=0.45\textwidth]{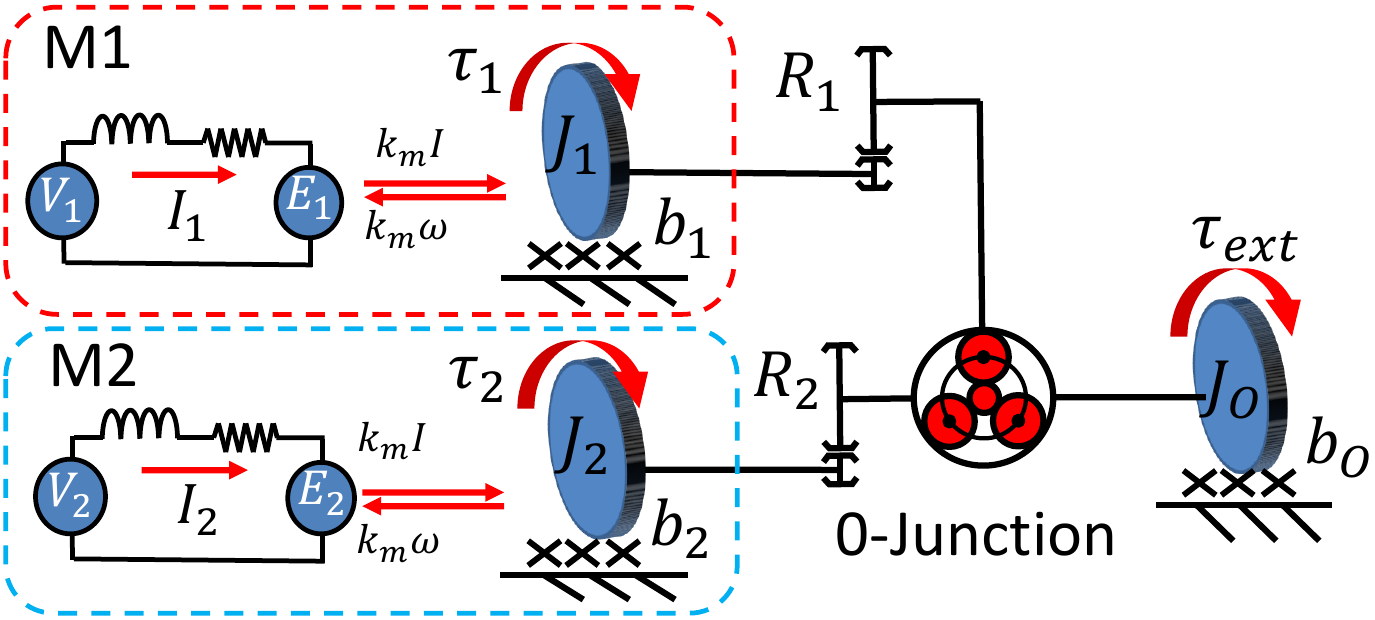}
	\caption{Lumped-parameter dynamic model of a DSDM}
	\label{fig:dynamics}
\end{figure}

\begin{figure}[htp]
	\centering
		\includegraphics[width=0.40\textwidth]{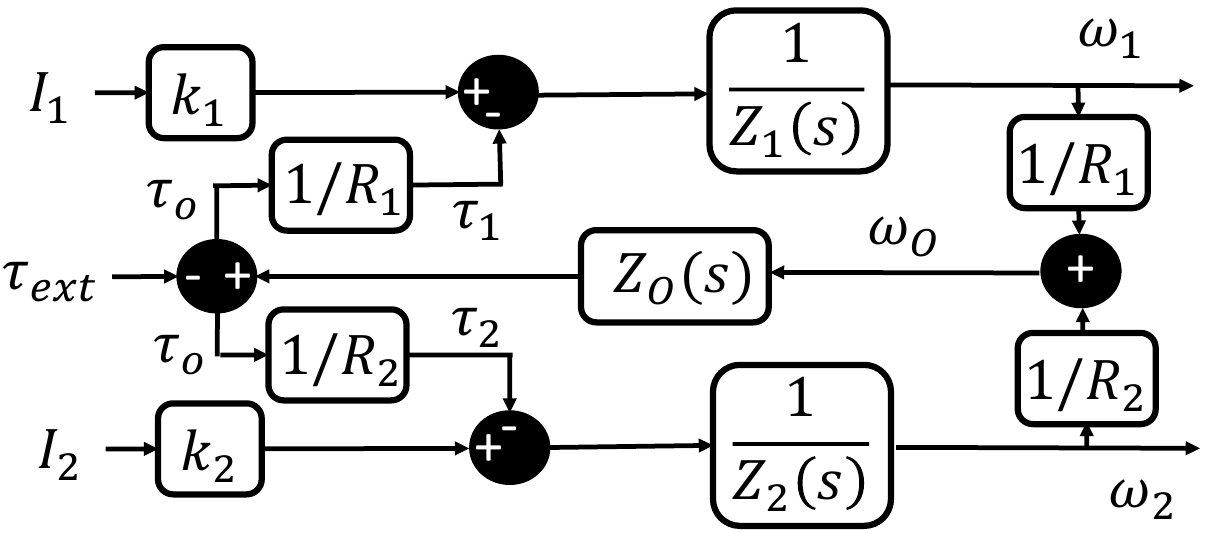}
	\caption{Dynamics of a DSDM}
	\label{fig:dynamics_block}
\end{figure}

Fig. \ref{fig:dynamics} shows a lumped-parameter dynamic model of a DSDM when the brake is open (high-speed mode). $J_i$ and $b_i$ are the inertia and damping of the respective i-th ports, $I_1$, $I_2$, $k_1$, $k_2$ are the currents and the torque constants of M1 and M2. Applying Newton's law on each ports yields the following equations of motions:
\begin{align}
\tau_{ext} - \tau_o &= Z_o(s) w_o \\
k_1 I_1    - \tau_1 &= Z_1(s) w_1 \\
k_2 I_2    - \tau_2 &= Z_2(s) w_2
\end{align}
where $Z_i(s) = J_i s + b_i$ represents the mechanical impedance of the i-th ports. Note that the system is coupled due to the constraint given by eq. \eqref{eq:kinematic} and \eqref{eq:torque}, and that there is only two degrees of freedom among the three ports. Fig. \ref{fig:dynamics_block}, illustrate the coupled equations motion in block diagram form.

It is then possible to eliminate one variable and express the dynamic as the following system of two equations:
\begin{align}
\left[
\begin{array}{l}
 k_1 I_1 \scriptstyle+\textstyle \frac{\tau_{ext}}{R_1} \\
 k_2 I_2 \scriptstyle+\textstyle \frac{\tau_{ext}}{R_2} \\
\end{array}
\right]
=
\left[
\begin{array}{c c}
\frac{Z_o(s)}{R_1}  &   \scriptstyle Z_1(s)   \\
\frac{Z_o(s)}{R_2}  \scriptstyle + R_2 Z_2(s) & \scriptstyle-\textstyle \frac{R_2}{R_1} \scriptstyle Z_2(s)   \\
\end{array}
\right]
\left[
\begin{array}{l}
w_o     \\
w_1     \\
\end{array}
\right]
\end{align}
The equations can be converted to state space form:
\begin{align}
\left[
\begin{array}{c}
\dot{w_o}\\
\dot{w_1}
\end{array}
\right] &= 
\frac{-1}{J_T}
\left[
\begin{array}{c c}
b_T &   0 \\
R_1 b_o  &  0 \\
\end{array}
\right]
\left[ \begin{array}{c}
w_o \\
w_1
\end{array} \right]+
\underline{B}
\left[ \begin{array}{c}
I_1 \\
I_2
\end{array} \right] 
\label{eq:ss}
\\
\text{with} \quad 
\underline{B} &= 
\frac{1}{J_T}
\left[
\begin{array}{c c}
k_1 R_1  &   k_2 R_1 \frac{R_1 J_1}{R_2 J_2}  \\
k_1(R_1^2+\frac{R_1^2 J_o}{R_2^2 J_2})  &  -k_2 \frac{R_1 J_o}{R_2 J_2} \\
\end{array}
\right] \\
 J_T &=  \scriptstyle \left[   J_o + R_1^2 J_1 + \left( \frac{R_1}{R_2} \right)^2 \frac{J_1}{J_2} J_o \right]\\
 b_T &= \scriptstyle \left[ b_o + \left( \frac{R_1}{R_2} \right)^2 \frac{J_1}{J_2} b_o \right] 
\end{align}
where the external torque and the damping at each input port are eliminated for brevity. From eq.\eqref{eq:ss} it can be seen that if $R_1<<R_2$ then M1 current $I_1$ dominates the output responses during high-speed mode.

%%%%%%%%%%%%%%%%%%%%%%%%%%%%%%%%%%%%%%%%%%%%%%%%%%%%%%%%%
\section{Control}
\label{sec:ctl}
%%%%%%%%%%%%%%%%%%%%%%%%%%%%%%%%%%%%%%%%%%%%%%%%%%%%%%%%%

%%%%%%%%%%%%%%%%%%%%%%%%%%%%%%%%%%%%%%%%%%%%%%%%%%%%%%%%%
\subsection{Output nullspace}
\label{sec:tran}

From the kinematic input-output view point, the DSDM actuator has one redundant degree of freedom. In other words, there is an infinite number of combinations of $w_1$ and $w_2$ producing the same output speed $w_0$, from eq.\eqref{eq:kinematic}:  
\begin{align}
\left[
w_o
\right] = 
\left[
\begin{array}{c c}
\frac{1}{R_1} & \frac{1}{R_2}
\end{array}
\right]
\left[
\begin{array}{c}
w_1 \\
w_2 \\
\end{array}
\right]
\end{align}
A vector perpendicular to the above coefficient vector forms the null space of the DSDM actuator system. Any input combination in this direction produces zero output speed:
\begin{align}
\left[
\begin{array}{c}
w_1 \\
w_2 \\
\end{array}
\right]=
\underbrace{\left[
\begin{array}{c}
1 \\
-R_2/R_1 \\
\end{array}
\right]}_{\text{Nullspace Projection}}
u  \; \rightarrow \;
w_0 = 0 \quad \forall u \in \Re
\label{eq:kinematicnullspace}
\end{align}
Interestingly, a similar expression can be obtained for the dynamics of the output in response to current inputs at the two motors, from eq.\eqref{eq:ss}:
\begin{align}
J_T \dot{w}_o +
b_T  w_o
=&
\left[ \begin{array}{c c}
k_1 R_1 & k_2 R_1 \frac{R_1 J_1}{R_2 J_2}
\end{array} \right]
\left[ \begin{array}{c}
I_1 \\
I_2
\end{array} \right]
\label{eq:output}
\end{align}
Hence, there is a one degree of freedom space of inputs $I_1$ and $I_2$ that do not affect the output:
\begin{align}
\left[ \begin{array}{c}
I_1 \\
I_2
\end{array} \right]
 = 
\underbrace{\left[ \begin{array}{c}
\frac{J_1}{k_1} \\
-\frac{R_2 J_2}{R_1 k_2} 
\end{array} \right]}_{\text{Nullspace Projection}} u
\; \rightarrow \;
J_T \dot{w}_o +
b_T  w_o = 0 \quad \forall u \in \Re
\label{eq:dyn_null_proj}
\end{align}

\subsection{High-force mode}
During high-force mode, with M1 locked by the brake, the actuator behaves like a regular geared EM motor and standard speed or position control scheme can be used.

\subsection{High-speed mode}

\begin{figure*}[tp]
	\centering
		\includegraphics[width=1.00\textwidth]{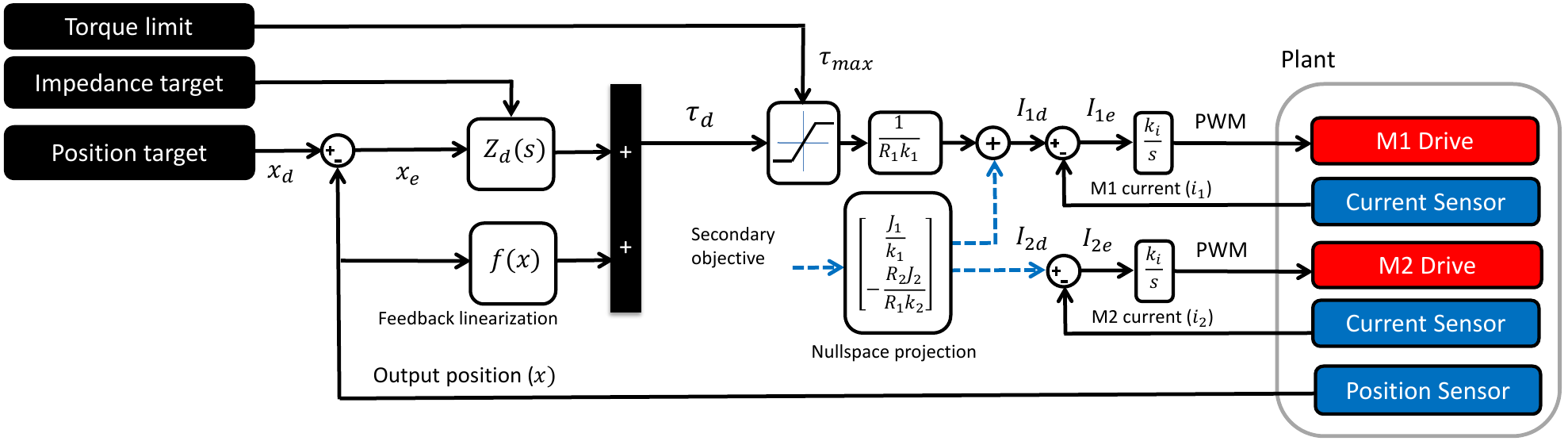}
	\caption{High-speed mode control loop}
	\label{fig:HS_loop}
\end{figure*}

With a small reduction ratio $R_1$ the DSDM actuator can be controlled like a direct drive motor: the output torque can be controlled directly by controlling M1 current, see inner feedback loop in Fig. \ref{fig:HS_loop}. The output loop can be used for various objectives. With the measurement of output position and velocity, impedance control, friction compensation and feedback linearization can be implemented. Moreover, exploiting the nullspace, a secondary objective can be brought into the system without influencing the output. In case the first motor is overloaded, for example, the second motor can reduce the load by projecting inputs through the null space, producing no effect upon the output, but changing the proportion of the two input commands. Note that the nullspace projection vector, see eq.\eqref{eq:dyn_null_proj}, depends only on parameters associated with the motors. Therefore, it is possible to project the secondary controller inputs on the output nullspace even if the output dynamic parameters that include the load inertia and damping are unknowns. The secondary controller can be used not only for load balancing but also for minimizing power consumption, avoiding speed saturation of M1, etc.

\subsection{Seamless mode transitions}

Unlike powertrains, where the load is mostly a large inertia and can be disengaged momentarily during gear shiftings, robotic actuators face a wide range of dynamic load, which must be controlled continuously even during a transition. For instance, if it drives a spring-like load, even a very short disengagement from the load might lead to large undesirable output motion. This section addresses transition control between the two modes.

\subsubsection{From high-force mode to high-speed mode}
This transition control is simple. The locking brake can be released anytime, M1 is then instantaneously freed and the controller can immediately switch to the high-speed control mode.

\subsubsection{From high-speed mode to high-force mode}
For this transition, M1 speed $w_1$ must be brought to zero so that the locking brake can be engaged. Two algorithms, a kinematic and a dynamic approach, can be considered. The former assumes local high gain velocity feedback controls for the individual motors. Hence velocities $w_1$ and $w_2$ can be treated as control inputs. Using the nullspace projection vector from eq.\eqref{eq:kinematicnullspace}, the kinematic control law can be written as 
\begin{align}
\left[
\begin{array}{c}
w_1 \\
w_2 \\
\end{array}
\right]=
\underbrace{\left[
\begin{array}{c}
1 \\
-R_2/R_1 \\
\end{array}
\right]}_{\text{Nullspace Projection}}
u_1 + 
\left[
\begin{array}{c}
0 \\
R_2 \\
\end{array}
\right] u_2
\label{eq:kinematicsys}
\end{align}
leading to
\begin{align}
w_o = u_2 \quad \text{with} \quad w_1 = u_1
\end{align}
Therefore during transitions, using $u_1$ velocity $w_1$ can be driven to zero, while fully controlling the output velocity using $u_2$. The kinematic control law is valid only when high fidelity velocity controls are available. Alternatively the dynamic control algorithm does not require this. While running the high-speed control scheme of Fig. \ref{fig:HS_loop}, a braking law for $w_1$ can be used in parallel as the secondary controller projected on the output nullspace:
\begin{align}
\left[ \begin{array}{c}
I_1 \\
I_2
\end{array} \right]
 = 
\underbrace{\left[ \begin{array}{c}
\frac{J_1}{k_1} \\
-\frac{R_2 J_2}{R_1 k_2} 
\end{array} \right]}_{\text{Nullspace Projection}} \underbrace{-C w_1}_{\text{Braking Law}} + 
\left[ \begin{array}{c}
\frac{1}{k_1 R_1} \\
0 
\end{array} \right]  \tau_d
\end{align}
leads  to:
\begin{align}
J_T \dot{w}_o +
b_T  w_o
=& \, \tau_d  \quad \text{and} \quad \dot{w}_1 = -C w_1 + f(w_o,\tau_d) %\scriptstyle \frac{R_1 b_o}{J_T} \textstyle w_o
\end{align}
Hence, the output is not influenced by the braking law due to orthogonality, and is still controlled using the desired torque $\tau_d$ determined by the high-speed mode controller. On the other hand, $w_1$ is directly influenced by the braking law but also by the output speed and the desired output torque $\tau_d$. Mathematically, it would be possible to also fully uncouple $\dot{w}_1$ equation, but the control law would not be practical in the scenario of $R_1<<R_2$ when considering torque and speed saturations. Increasing the gain $C$ will lead to faster braking of $w_1$, however $I_2$ will saturate if the gain is too large. A large $C$ can still be used for faster braking at a cost of deviation from the desired torque $\tau_d$. There is a trade-off, passed the $I_2$ saturation point, between fast braking of $w_1$ for fast transition and high-fidelity output torque control.

%%%%%%%%%%%%%%%%%%%%%%%%%%%%%%%%%%%%%%%%%%%%%%%%%%%%%%%%%
\subsection{Automatic mode selection}
\label{sec:mode}

\begin{figure}[H]
	\centering
		\includegraphics[width=0.45\textwidth]{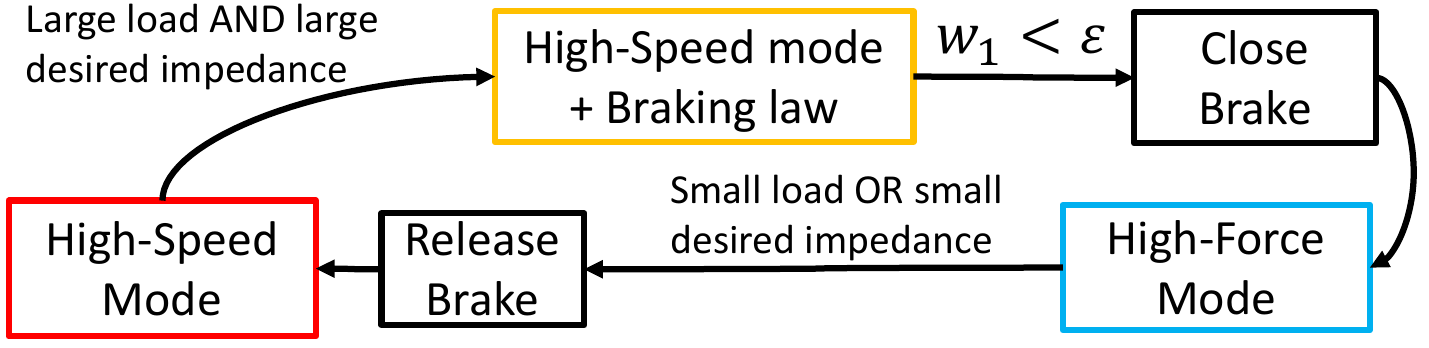}
	\caption{State machine for automatic mode selection}
	\label{fig:automaticflow}
\end{figure}

To simplify the programming of a DSDM actuator by an end-user, it is interesting to integrate an automatic mode selection in the control scheme. As shown in Fig. \ref{fig:automaticflow}, the mode selection could be made based the on the load conditions and a user-provided desired impedance for each task. When a large impedance is desired, the actuator automatically select the mode that allow the fastest convergence to the target given the load conditions. Hence, the actuator will use high-speed mode, except when encountering resistance it will automatically "down shift" to high-force mode. If a low impedance or torque limitation is desired, then the DSDM actuator must stay in high-speed mode to enforce those specifications. 
%

%%%%%%%%%%%%%%%%%%%%%%%%%%%%%%%%%%%%%%%%%%%%%%%%%%%%%%%%%%%%%%%%%%%%%%%%%%%%%

\section{EXPERIMENTAL VALIDATIONS}
\label{ev}

%%%%%%%%%%%%%%%%%%%%%%%%%%%%%%%%%%%%
\subsection{Test bench design}

\begin{figure}[htb]
	\centering
		\includegraphics[width=0.48\textwidth]{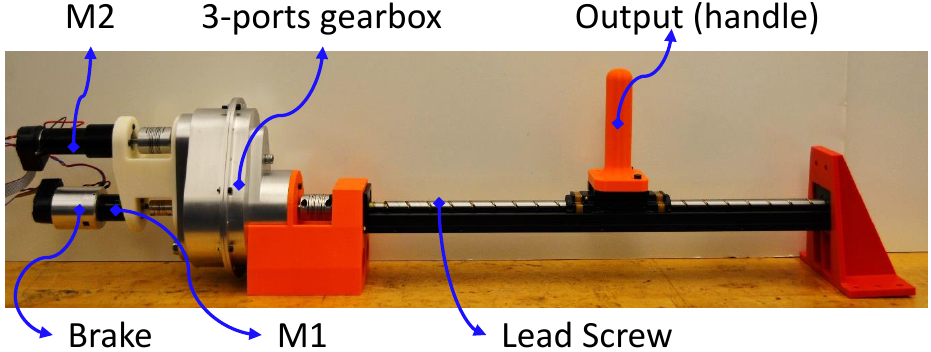}
	\caption{Proof-of-concept DSDM linear actuator}
	\label{fig:proto_linear}
\end{figure}

\begin{figure}[htb]
        \centering
				\subfloat[Prototype]{
				\label{fig:gears}
				\includegraphics[width=0.20\textwidth]{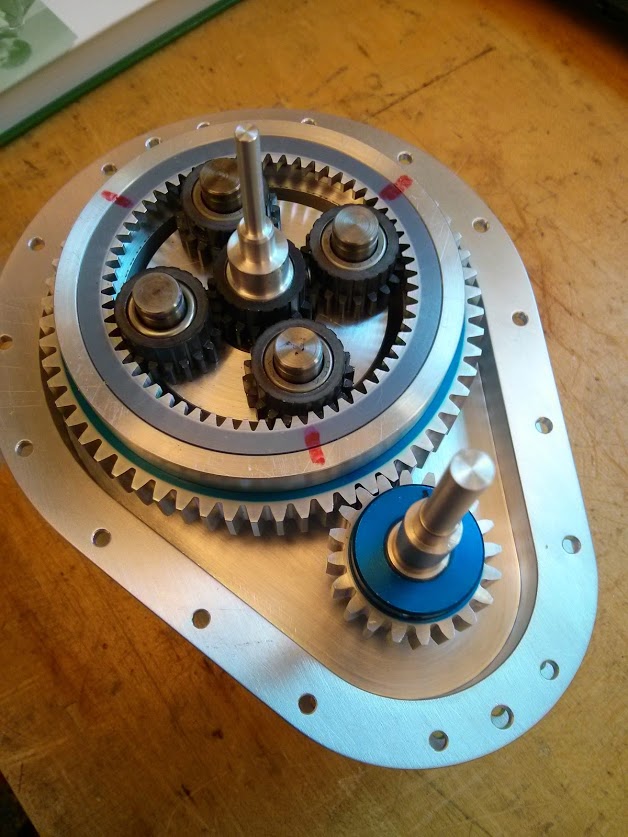}
				\label{fig:gearsreal}}
        \subfloat[CAD section view]{
				\includegraphics[width=0.20\textwidth]{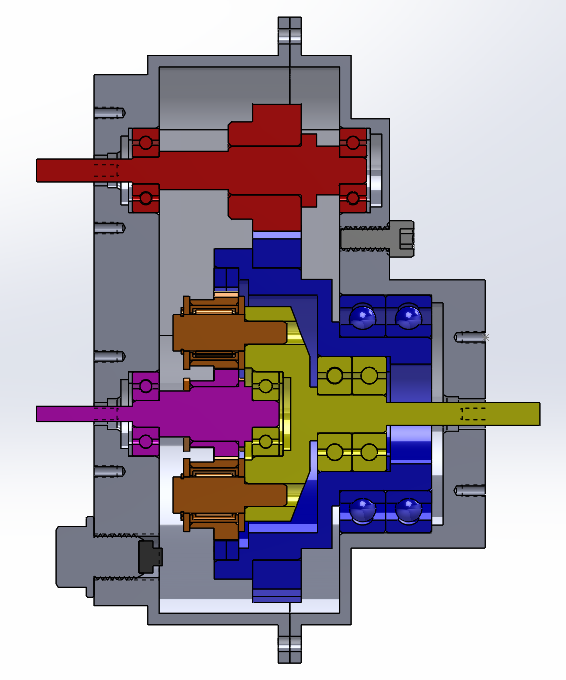}
				\label{fig:gearscad}}
        \caption{Custom 3-ports gearbox}\label{fig:gears2}
\end{figure}

A test bench was developed to demonstrate the functionality of the DSDM concept, see Fig. \ref{fig:proto_linear}. Discrete components (motors, brake, gearhead and differential) are used for the ease of implementation and future modifications. Two 20 Watts DC motor (Maxon RE 25) are used for both M1 and M2 positions. M1 is equipped with an electromechanical brake (Maxon AB 28) and M2 with a gearhead reductor of 18:1 (Maxon GP 32). The differential is a custom made 3-ports planetary gear, see Fig. \ref{fig:gears2}, with an additional reduction of 3:1 on the ring gear input. The planetary gearing is designed with a ratio $N$ of 3 (see eq.\eqref{eq:kinematic}), hence $R_1$= 4 and $R_2$= 72. The output is connected to a 20 mm lead ballscrew, leading to specifications given by Table II.
\begin{table}[htb]
	\centering
		\label{tab:proto}
		\begin{tabular}{ c c c c}
		\hline
		Mode                &  Max Force  & Max Speed         & Apparent Output Mass      \\
		\hline\hline
		High-force          & 600 N       &  40 mm/sec        & 750 kg       \\
		High-speed          & 30  N       &  0.7 m/sec        & 2   kg      \\
		\hline			\\
		\end{tabular}
			\caption{Prototype specifications using two 20 W motors}
		\vspace{-5pt}
\end{table}
%

%%%%%%%%%%%%%%%%%%%%%%%%%%%%%%%%%%%%
\subsection{Controller Implementations}

The controller was implemented using a \textit{NI Compact Rio} platform and two \textit{NI 9505} DC motor drives. All inner current control loops are implemented on the FPGA at a 20 kHz sampling rate and the outer control loops are implemented in the real time OS at a 500 Hz sampling rate. Two encoders provide the position at both motor shafts and the output position is computed from those values using the kinematic relation of eq.\eqref{eq:kinematic}. The current is sensed with a Hall effect sensor (ACS712) and the speed data is computed using filtered discrete differentiation of the position data.

%%%%%%%%%%%%%%%%%%%%%%%%%%%%%%%%%%%%
\subsection{High-force mode}

\begin{figure}[htb]
	\centering
		\includegraphics[width=0.45\textwidth]{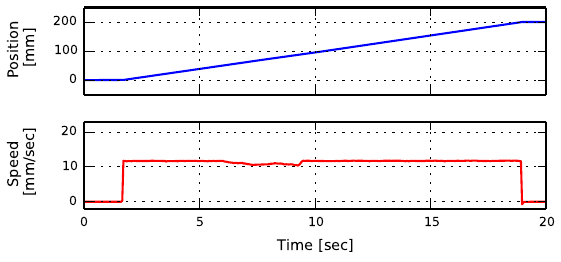}
	\caption{Output motion using high-force mode for a 200 mm target step}
	\label{fig:HF_plot}
\end{figure}

For the high-force mode experiment, M2 is controlled using an experimentally tuned PI with an anti-windup scheme. Fig. \ref{fig:HF_plot} shows the response to a 200 mm step of the reference position, with a perturbation at $t\,$= 5 sec (human applying about 50 lbs), see experiment \#1 in the video. The results show that the actuator is speed limited (maximum supply voltage reached) for most of the course, but very strong as illustrated by a small speed loss during the perturbation. The actuator precision is also very good as the final positioning error is less than a micron (neglecting the backlash as the position sensor is upstream of the gearbox). Hence, the actuator acts like a strong and precise displacement source as desired in this mode.

%%%%%%%%%%%%%%%%%%%%%%%%%%%%%%%%%%%%
\subsection{High-speed mode}

\begin{figure}[htb]
	\centering
		\includegraphics[width=0.45\textwidth]{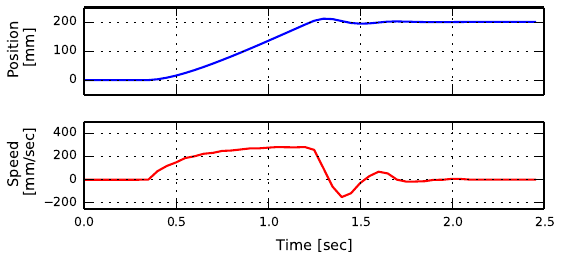}
	\caption{Output motion using high-speed mode for a 200 mm target step}
	\label{fig:HS_plot}
\end{figure}

For the high-speed mode experiment, the inner current controllers are pure integrator leading to closed-loop bandwidth well over 500 Hz. Part of the prototype coulomb friction (mainly in the ballscrew) is canceled using the feedback linearization loop, see Fig. \ref{fig:HS_loop}, and the prescribed impedance is a pure stiffness, relying on the natural friction of the system to provide the damping. Fig. \ref{fig:HS_plot} shows the results of, as before, the response to a 200 mm reference position step, see experiment \#2 in the video. The result with the high-speed mode is much more dynamic than the high-force mode response, and a maximum speed of 0.3 m/sec was reach during the motion. The actuator overshoot a little bit before stabilizing on the target. Also, when the desired impedance is set to zero, the actuator is easy to back-drive by hand, as illustrated by experiment \#3 in the video. Moreover, the actuator can limit its output force and handle collision without stability issues as demonstrated in experiment \#5 in the video. An inherent advantage of not relying on force feedback to control the impedance.
%

%%%%%%%%%%%%%%%%%%%%%%%%%%%%%%%%%%%%
\subsection{Mode transitions}

\begin{figure}[htb]
	\centering
		\includegraphics[width=0.45\textwidth]{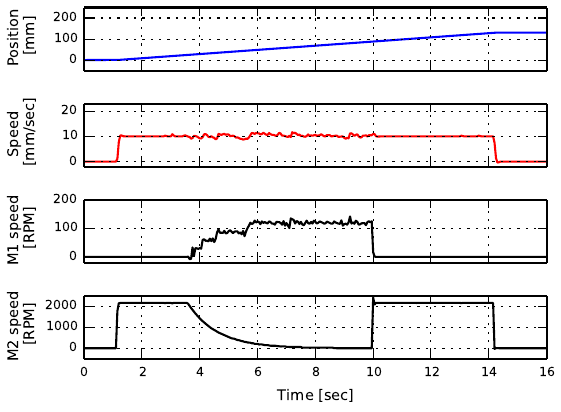}
	\caption{Mode transitions with a constant output speed target}
	\label{fig:shifts_plots}
\end{figure}

Fig. \ref{fig:shifts_plots} shows the results from an experiment were the actuator first control the output speed using high-force mode, then switch to high-speed mode (M2 speed is then brought to zero for demonstration purposes) and then switch back to high-force mode, see experiment \#4 in the video. The results show that the transitions are seamless and that the output is always under control, here maintaining a constant speed. Since the actuator is fighting a dissipative load (the friction in the ballscrew), a constant force must be applied at all time, if not the output would suddenly stop. Note, the noisy behavior when using high-speed mode at low speed is due to a limited angular velocity resolution on $w_1$. 

\begin{figure}[htb]
	\centering
		\includegraphics[width=0.45\textwidth]{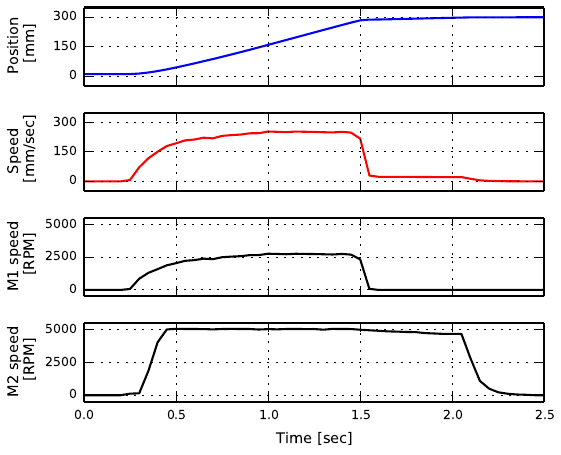}
	\caption{Automatic down shift during contact with a spring-like load}
	\label{fig:auto_shift}
\end{figure}

Fig. \ref{fig:auto_shift} shows the results from an experiment were the actuator is controlled in position (target at 300 mm) and automatically select the mode as proposed in section \ref{sec:mode}. Here the actuator first accelerate quickly using high-speed mode since there is no resistance and switch back to high-force mode after impacting a ball to continue its course (experiment \#6 in the video). Hence the transition is done while in contact with a spring-like load. The results show that the DSDM is still able to switch mode quickly even during this dynamic scenario. This experiment also illustrate how a single task can take advantage of the two operating modes of the actuator; fast reaching and large force as soon as the contact is made.

%%%%%%%%%%%%%%%%%%%%%%%%%%%%%%%%%%%%%%%%%%%%%%%%%%%%%%%%%%%%%%%%%%%%%%%%%%%%%
\section{CONCLUSION AND OUTREACH}

In this paper, the use of distinctively different gear ratios with EM motors in robotics is proposed. This allow for an effective use of power over a wide range of torque-speed conditions, enabling smaller and more efficient actuator systems. Furthermore, very different gear ratios lead to drastic intrinsic impedance change, enabling a displacement-source type of behavior with the large non-back-drivable reduction ratio and good force control capability with the small reduction ratio. Additionally, a novel dual-motor implementation of dual-speed is proposed to address the problematics of gear shifting in the context of robotics. The proposed DSDM actuator can keep full control of the output during gear shifting, with a controller exploiting the redundancy of inputs. The DSDM actuator is advantageous for many actual robotic applications, but also enabled new applications that are impossible with regular actuators because the requirements are too conflicting to meet with a single motor of usable size. 

Also, many aspect of using multiple gear ratios in a robotic context have yet to be explored. Gear shifting is a very non-linear process and it would be interesting to explore with greater depth how it should be integrated in a robot routine. Regarding the DSDM concept, in this first iteration two identical motors were used. However, it could be more effective to use asymmetric motors optimized for their particular roles. For instance M1 could be optimized for force bandwidth and M2 for power density and precision. Generalizing, it would even be possible to use different technologies for the force source and the displacement source, recall Fig. \ref{fig:lever}.

%For instance, as illustrated on Fig. , a mobile robot capable of dynamic locomotion and CNC-like manufacturing operations. 
%
%
%\begin{figure}[htbp]
	%\centering
		%\includegraphics[width=0.35\textwidth]{../img/robot.JPG}
	%\caption{aaa}
	%\label{fig:robot}
%\end{figure}

%%%%%%%%%%%%%%%%%%%%%%%%%%%%%%%%%%%%%%%%%%%%%%%%%%%%%%%%%%%%%

\section*{ACKNOWLEDGMENTS}
Special thanks to Lluis Penalver-Aguila and James Torres for the help with the detailed drawing and the manufacturing of the gearbox.

%%%%%%%%%%%%%%%%%%%%%%%%%%%%%%%%%%%%%%%%%%%%%%%%%%%%%%%%%%%%%%

%********************************************************************************
% References                                                                     
%*************************************************************************